\documentclass[runningheads]{llncs}
\usepackage{array}
\usepackage{xspace}
\usepackage{amsmath,amssymb,amsfonts}
\usepackage{graphicx}

\newcommand{\tup}[1]{\ensuremath{\langle #1 \rangle}\xspace}

\newcommand{\A}{\ensuremath{\mathcal{A}}\xspace}
\newcommand{\C}{\ensuremath{\mathcal{C}}\xspace}
\newcommand{\D}{\ensuremath{\mathcal{D}}\xspace}

\newcommand{\I}{\ensuremath{\mathcal{I}}\xspace}

\newcommand{\N}{\ensuremath{\mathrm{N}}\xspace}

\newcommand{\tmS}{\ensuremath{\mathcal{S}}\xspace}
\newcommand{\T}{\ensuremath{\mathcal{T}}\xspace}

\newcommand{\DL}{\ensuremath{\mathcal{DL}}\xspace}
\newcommand{\DLN}{\ensuremath{\mathcal{DL}^\N}\xspace}
\newcommand{\ALC}{\ensuremath{\mathcal{ALC}}\xspace}
\newcommand{\EL}{\ensuremath{\mathcal{EL}}\xspace}
\newcommand{\ELP}{\ensuremath{\mathcal{EL}^{++}}\xspace}

\newcommand{\KB}{\ensuremath{\mathcal{KB}}\xspace}

\newcommand{\NC}{\ensuremath{\mathsf{N_C}}\xspace}

\newcommand{\NI}{\ensuremath{\mathsf{N_I}}\xspace}
\newcommand{\NR}{\ensuremath{\mathsf{N_R}}\xspace}
\newcommand{\pre}{\mathsf{pre}}
\newcommand{\con}{\mathsf{con}}

\newcommand{\nent}{\ensuremath{\mathrel{\mbox{$|\hspace*{-.48em}\approx$}}}\xspace}

\newcommand{\PT}{\ensuremath{\mathrm{P}}\xspace}

\newcommand{\infrule}[2]{$
  \renewcommand{\arraystretch}{1}
  \begin{array}{c}
    #1 \\ \hline #2
  \end{array}
  $}
  
\newcommand{\hide}[1]{}
\newcommand{\ty}{\ensuremath{\mathbf{T}}\xspace}

\begin{document}
\title{Defeasible reasoning in Description Logics: an overview on  \DLN}

\author{Piero A. Bonatti\inst{1} 
\and Iliana M. Petrova\inst{1}
\and Luigi Sauro\inst{1}
}

\institute{Dep. of Electrical Engineering and Information Technologies, Universit\`a di Napoli Federico II, Italy}

\maketitle

\begin{abstract}
\DLN is a recent approach that extends description logics with defeasible reasoning capabilities. 
In this paper we provide an overview on 
\DLN, illustrating the underlying knowledge engineering requirements as well as the characteristic features that preserve \DLN from some recurrent semantic and computational drawbacks.  
We  also compare  \DLN with some alternative nonmonotonic semantics, enlightening the relationships between the KLM postulates and \DLN.
\end{abstract}

\section{Introduction}\label{sec:intro}

In complex areas  such as law and science, knowledge has been in centuries formulated by primarily describing  prototypical instances and properties, and then by \emph{overriding} the general theory  to include possible exceptions. 
For example, many laws are formulated by adding new norms that, in case of conflicts, may   partially or completely override the previous ones. 
Similarly, biologists have been incrementally introducing exceptions to general properties. For instance, the human heart is usually located in the left-hand half of the thorax. Still there are exceptional individuals, with so-called \emph{situs inversus}, whose heart is located on the opposite side. Eukariotic cells are those with a proper nucleus, by definition. Still they comprise mammalian red blood cells, that in their mature stage have no nucleus.\footnote{All of these examples are introduced and discussed in \cite{DBLP:conf/psb/Rector04,DBLP:journals/ijmms/StevensAWSDHR07}.}

Also many modern applications and methodologies in Computer Science rely on some sort of overriding mechanism. In Object Oriented Programming the definitions in a subclass may override
any conflicting bindings belonging to its superclasses.
Analogously, formal languages designed to describe role-based access control or other privacy policies generally allow 
to formulate  default conditions, such as \emph{open} and
\emph{closed} policies,%
\footnote{ If no explicit authorization has been specified for a given access request, then an open policy permits the access while a closed policy denies it.  } conflict resolution methods such as
\emph{denials take precedence}, and authorization inheritance with
exceptions \cite{DBLP:conf/dagstuhl/BonattiS03}. 

Summarizing, the mentioned fields manifest to a large extent different forms of defeasible knowledge where general axioms can be recanted in special cases by employing some suitable overriding mechanism. 
Nevertheless, this natural approach cannot be directly adopted in designing Semantic Web ontologies. In fact, the underlying descriptions logics (DLs), which are based on the  monotonic semantics of FOL, do not allow to express and reason on defeasible knowledge and exceptions. Consequently, several authors advocated nonmonotonic logics as a useful means to address this limitation and proposed different formalisms based on circumscription \cite{DBLP:conf/aaai/BonattiFS11,DBLP:conf/semweb/BonattiFS10,DBLP:journals/jair/BonattiFS11}, autoepistemic logic \cite{donini97autoepistemic,DBLP:journals/tocl/DoniniNR02}, typicality operators \cite{DBLP:conf/jelia/GiordanoGOP08,DBLP:journals/ai/GiordanoGOP13,DBLP:journals/fuin/GiordanoOGP09}, or rational closure \cite{DBLP:journals/ai/GiordanoGOP15,DBLP:conf/jelia/CasiniS10,DBLP:journals/ai/Bonatti19}, just to mention a few.   

In this context, \DLN \cite{DBLP:journals/ai/BonattiFPS15,DBLP:journals/ai/BonattiS17,DBLP:conf/semweb/BonattiPS15} is a recent family of nonmonotonic  DL specifically designed to  meet the knowledge engineering requirements that come from the aforesaid application domains.
\DLN  is prototype oriented: it uses
 $\N C$ to denote the normal/prototypical instances of a concept $C$, and extends terminological axioms with prioritized \emph{defeasible inclusions} (DIs) $C\sqsubseteq_n D$. 

A difficulty arising at a design level is that the notion itself of prototype can be inherently ambiguous. Prototypes may represent in a \emph{frequentistic} fashion the properties that are
shared by the majority of the instances, or they can differently be  interpreted \emph{idealistically} as platonic models that might not exist in the real world due to their degree of perfection.     
 \DLN does not aim at encompassing all different, and philosophically interesting,  notions of prototype; 
being plainly application-oriented, it is rather inspired by what McCarthy calls \emph{communication and database
  storage conventions} \cite{McC86}.
In this perspective,
a prototype $\N C$ is meant to factorize the common features of
the concept $C$ and confine exceptional subclasses to an explicit detailed axiomatization (so as to reduce the
size and cost of knowledge bases and improve their
readability). Thus, defeasible inclusions $C\sqsubseteq_n D$ mean (roughly speaking): ``\emph{by default, all prototypical instances that satisfy $C$ satisfy also $D$, unless stated otherwise}'', that is, unless some higher priority axioms contradict this implication. If such a contradiction arises, then $C\sqsubseteq_n D$ is \emph{overridden}. The standard/prototypical instances of $C$ are required to satisfy all the DIs that are not overridden in $C$.

As mentioned above, also other nonmonotonic logics support defeasible inheritance with overriding  in general.
Nevertheless, in each of these previous approaches, either some desiderable features are missing or some natural inferences do not hold. Moreover, they are generally based on complex semantics which make defeasible reasoning difficult to track.\footnote{For example,  circumscription identifies, in case of conflicting nonmonotonic axioms,  all optimal repairs and then computes the inferences that hold for all repairs.} 
In this respect,   DLN's behavior is easier to grasp, and is expected to facilitate knowledge engineers in formulating and validate ontologies at a large scale, while producing the expected conclusions.

Finally, apart from rational closure and restricted forms of typicality \cite{DBLP:conf/dlog/0001GPR17}, defeasible reasoning significantly increases the computational complexity of standard reasoning tasks even in low-complexity description logics \cite{DBLP:conf/lpnmr/GiordanoGOP09,DBLP:conf/ijcai/BonattiFS09,DBLP:journals/jair/BonattiLW09}. Conversely, in all DL fragments of pratical interest, \DLN does not manifest a higher complexity with respect to the classical counterpart. Moreover, efficiency can be further enhanced through a range of optimization techniques, including  modularization \cite{DBLP:conf/semweb/BonattiPS15}.

This paper is meant to illustrate \DLN and its main features, extending previous discussions of \DLN's properties with some recent contributions to rational closure.
The paper is organized as follows. In the next section we briefly recall the basics of monotonic description logics.  In Section~\ref{sec:DLN} we introduce \DLN and provide a few examples of knowledge bases and inference. Section~\ref{sec:feat} compares \DLN with the other major nonmonotonic DLs in terms of practical engineering requirements, and in terms of logical properties, centred around the KLM postulates. The paper is concluded by a summary and a list of interesting topics for further work. 

\section{Preliminaries}\label{sec:pre}

Description logics are a family of  formal languages representing the logical foundations 
of the W3C Ontology Web Language (OWL2). 
They offer a variegated set of \emph{logical constructors} and \emph{axioms} that balance between 
expressiveness and  computational complexity according to the application needs. 
Due to space limitations,  we refer to \cite{Baader-et-al-03b} for a comprehensive overview. 
Here, we just introduce the DL fragment \ALC, which allows to understand the examples that will follow.

An alphabet or signature consists of  a set $\NC$ of \emph{concept
  names}, a set $\NR$ of \emph{role names}, and (possibly) a set $\NI$
of \emph{individual names} (all countably infinite). 
Thereafter, metavariables $A$, $B$ will range over concept names, $R$ and $S$ over roles, and $a$, $b$ and
$d$ over individual names. The term
\emph{predicate} will refer to a generic element of $\NC \cup \NR$.

In DLs, a wide range of operators allow to inductively formulate \emph{compound concepts}. 
The logic \ALC, in particular,  compound concepts are defined by the following grammar:
\[
C,D ::= A \mid \top \mid \bot \mid \neg C \mid C \sqcap D \mid C \sqcup
D \mid \exists R.C \mid \forall R.C \,.
\]
Note, however,  that our framework applies also to more expressive DLs such as $\mathcal{SROIQ(D)}$ that constitutes the foundation of the full standard OWL2.

%
\begin{figure}[tb]
  \begin{center}
    \small
    \leavevmode
    \begin{tabular}{|l|c|l|}
     \hline Name &Syntax&Semantics\\ \hline\hline & &
      \\[-1em]
      top & $\top$ & $\Delta^\I$ \\ \hline & &\\[-1em]
      bottom & $\bot$ & $\emptyset$ \\ \hline & &\\[-1em]
      negation&$\neg C$&$\Delta^\I \setminus C^\I$\\ \hline & &\\[-1em]
      conjunction&$C\sqcap D$&$C^\I\cap D^\I$\\ \hline & &\\[-1em]
      disjunction&$C\sqcup D$&$C^\I\cup D^\I$\\ \hline & &\\[-1em]
       $\exists$
       restriction
      &$\exists R. C$& 
      $\{ d \in \Delta^\I \mid  \exists e\in\Delta^\I.[ (d,e) \in R^\I \land e\in\C^\I] \}$\\ \hline & &\\[-1em]
       $\forall$
       restriction
      &$\forall R. C$& 
      $\{ d \in \Delta^\I \mid  \forall e\in\Delta^\I.[ (d,e) \in R^\I \rightarrow e\in\C^\I] \}$ 
      \\ \hline 
       \end{tabular} 
    \caption{Syntax and semantics of some common constructs.}
    \label{tab:syntax-semantics}
  \end{center}
\end{figure}

The semantics of DLs is defined in terms of
\emph{interpretations} $\I=\tup{\Delta^\I,\cdot^\I}$. The \emph{domain}
$\Delta^\I$ is a non-empty set of individuals and the
\emph{interpretation function} $\cdot^\I$ maps each concept name
$A\in\NC$ to a subset $A^\I$ of $\Delta^\I$, each role name
$R\in\NR$ to a binary relation $R^\I$ on $\Delta^\I$, and each
individual name $a \in \NI$ to an individual $a^\I \in \Delta^\I$.
The extension of $\cdot^\I$ to \ALC compound concepts
is inductively defined as shown in the third column of
Figure~\ref{tab:syntax-semantics}. An interpretation \I is called a
\emph{model} of a concept $C$ if $C^\I \neq \emptyset$. 

A \emph{(general) TBox} is a finite set of \emph{concept inclusions
  (CIs)} $C \sqsubseteq D$. As usual, we use $C \equiv D$ as an
abbreviation for $C \sqsubseteq D$ and $D \sqsubseteq C$. An
\emph{ABox} is a finite set of \emph{concept assertions} $C(a)$ and
\emph{role assertions} $R(a,b)$.  An interpretation \I
\emph{satisfies} (i) a CI $C \sqsubseteq D$ if $C^\I \subseteq D^\I$,
(ii) an assertion $C(a)$ if $a^\I \in C^\I$, and (iii) an assertion
$R(a,b)$ if $(a^\I,b^\I) \in R^\I$.  Then, \I is a (classical)
\emph{model} of a TBox \T (resp.\ an ABox \A) if \I satisfies all the
members of \T (resp.\ \A).

In this paper, we will sometimes mention some important DLs that have been
extensively studied in the literature and constitute the foundation of
semantic web standards. 
The logic $\EL$ supports only $\top$, $\sqcap$, and $\exists$. Its
extension $\EL^\bot$ supports also $\bot$.  The logic $\EL^{++}$ further
adds \emph{concrete domains} and some expressive role inclusions (see
\cite{DBLP:conf/ijcai/BaaderBL05} for further details). 

The logic DL-lite$_R$ \cite{DL-lite05} supports inclusions shaped like
$C\sqsubseteq D$ and $C\sqsubseteq \neg D$, where $C$ and $D$ range
over concept names and \emph{unqualified existential restrictions}
such as $\exists R$ and $\exists R^-$ (where $R^-$ is the inverse of role $R$). $\EL^{++}$ and
DL-lite$_R$, respectively, constitute the foundation of the OWL2
profiles OWL2-EL and OWL2-QL.  Both play an important role in
applications; their inference problems are tractable (the same holds
for some extensions of DL-lite$_R$, see
\cite{DBLP:journals/jair/ArtaleCKZ09}).


Finally, we will use in Section~\ref{sec:klm} boolean combinations of assertions and inclusions. Although these axioms are not directly allowed in DLs, they can be  simulated in $\mathcal{SROIQ(D)}$ through   
the universal role $U$. For example, $\neg (C\sqsubseteq D)$ and $(C_1\sqsubseteq D_1)\vee (C_2\sqsubseteq D_2)$ can be expressed as $\top\sqsubseteq \exists U.(C\sqcap \neg D)$ and $\top\sqsubseteq (\forall U.(\neg C_1\sqcup D_1)) \sqcup (\forall U.(\neg C_2\sqcup D_2))$, respectively.

\section{The defeasible logic \DLN}
\label{sec:DLN}

Given a classical description logic language \DL, let 
 \DLN be the
extension of \DL with a new concept name $\N C$ for each \DL concept $C$. $\N C$ is called a  \emph{normality concept} and denotes  the \emph{normal instances} of $C$.

\noindent
A  \DLN knowledge base is a disjoint union $\KB=\tmS\cup \D$ such that 
\begin{itemize}
\item $\tmS$ is a finite set of  \DL concept inclusions and assertions; 
\item $\D$ is a finite set of defeasible inclusions (DIs, for short)  $C\sqsubseteq_n D$ where $C$ and $D$ are \DLN concepts. 
\end{itemize}

\noindent
Thereafter, given a DI $\delta=C\sqsubseteq_n D$, by $\pre(\delta)$ and $\con(\delta)$ we denote $C$ and $D$, respectively.
A knowledge base is \emph{canonical} if $\pre(\delta)$ does not contain normality concepts, for all  $\delta\in\D$.

Roughly, $C\sqsubseteq_n D$ means: ``\emph{the normal instances of $C$ are instances of $D$, unless stated otherwise by  some higher priority axioms}''. As mentioned in the introduction DIs have an utilitarian purpose. They are meant to factorize the common properties that hold for normal entities, so as to minimize the amount of knowledge that must be explicitly encoded.

Defeasible inclusions are prioritized by a \emph{strict partial order} $\prec$ over $\D$. The intended meaning of $\delta_1\prec\delta_2$ is that $\delta_1$ has higher priority than $\delta_2$ and, in case of conflicts, it is preferable to sacrifice 
$\delta_2$. \DLN solves automatically only the conflicts that can be settled using $\prec$. Any other conflict shall be resolved by the knowledge engineer (typically by adding specific DIs).
Here, we focus on a priority relation which is determined by
so-called \emph{specificity}. Roughly speaking, specificity states that, in case of conflicts, the specific properties of $\pre(\delta_1)$
 override those of the more general concept $\pre(\delta_2)$:
\begin{equation}
  \label{specificity}
 \delta_1\prec\delta_2 \mbox{ iff }
  \tmS\models \pre(\delta_1)\sqsubseteq \pre(\delta_2) \mbox{ and }
  \tmS\not\models\pre(\delta_2)\sqsubseteq \pre(\delta_1) \,.  
\end{equation}
Note that \DLN is largely parametric with respect to which priority relation is used. An alternative choice could be, for example,  a priority relation   based on the ranking function  of rational closure adopted in \cite{DBLP:conf/jelia/CasiniS10}.

Due to space limitations, we refer to \cite{DBLP:journals/ai/BonattiFPS15} for the model-theoretic semantics of \DLN. Here, we present only its reduction to classical reasoning.

Let $\KB=\tmS \cup \D$ be a \DLN knowledge base  and $\alpha$ a query of interest, that can be either a CI or an assertion.
By $\KB \nent \alpha$ we mean that $\alpha$ is a \DLN \emph{semantic consequence} of $\KB$. 

\noindent
The classical reduction of $\nent$ requires some preliminary notions:
\begin{itemize}
\item  For all
DIs $\delta\in \D$ and all normality concepts $\N C\in\Sigma$, let 
$$
\delta^{\N C} = \big(\N C \sqcap \pre(\delta) \sqsubseteq \con(\delta)\big)\, ;
$$
\item for all sets of \DL axioms $\tmS'$ and all DIs $\delta$, let $\tmS'\downarrow_{\prec \delta}$ denote the result of removing from $\tmS'$ all the axioms $\delta_0^{\N C}$ such that $\delta_0$'s priority is not higher than $\delta$'s:
\[
\tmS'\downarrow_{\prec \delta} = \tmS' \setminus
     \{ \delta_0^{\N C} \mid \N C\in \Sigma \land
        \delta_0 \not\prec \delta \} \,;
\]
\item finally, let  $\delta_1,\ldots,\delta_{\vert \D\vert}$ be an arbitrary \emph{linearization} of $(\D,\prec)$, which means that $\{\delta_1,\ldots,\delta_{\vert \D\vert}\}=\D$ and for all $i,j=1,\ldots,\vert \D\vert$, if $\delta_i \prec \delta_j$ then $i<j$.
\end{itemize}

\noindent
Then, $\KB \nent \alpha$ holds iff $\KB^\Sigma\models \alpha$, where       
$\Sigma$ is the set of normality
concepts occurring in both $\KB$ and  $\alpha$, and    
$\KB^\Sigma$ is the classical knowledge base resulting from 
 the following inductive construction (where $i=1,2,\ldots,\vert \D\vert$):
\begin{eqnarray}
  \tmS^\Sigma_0 &=& \tmS \cup \big\{\N C\sqsubseteq C \mid \N C\in\Sigma\big\} 
  \\
  \label{KB-Sigma-constr-2}
  \tmS^\Sigma_i &=& \tmS^\Sigma_{i-1} \cup \big\{\delta_i^{\N C} \mid \N C\in\Sigma 
       \mbox{ and } \tmS^\Sigma_{i-1}\downarrow_{\prec \delta_i} \cup \{\delta_i^{\N C}\} \not\models
       \N C\sqsubseteq \bot \big\} \\
       \KB^\Sigma&=& \tmS^\Sigma_{\vert \D\vert}\,.
\end{eqnarray}

\noindent
In informal terms, the first step 
extends \tmS with the axioms $\N C\sqsubseteq C$ stating that the normal instances of $C$ are a fortiori instances of $C$. The construction proceeds by
processing the DIs $\delta_i\in\D$ in decreasing priority order; if
adding $\delta_i$ to the (higher priority) $\delta_j\prec\delta_i$
that have been previously selected does not make $\N C$ inconsistent,
as stated by (\ref{KB-Sigma-constr-2}), then $\delta_i^{\N C}$ is
included in $\KB^\Sigma$, otherwise $\delta_i^{\N C}$ is discarded
(overridden).


%


%

\begin{example}
  \label{situs-inversus-00.0}
  Recall that \emph{situs inversus} refers to humans whose heart is on the right-hand side of the thorax, differently from typical
  humans whose heart is on the opposite side.  If we stipulate that no heart can be simultaneously located on both sides, then a simple  axiomatization  is:\\  
\vspace*{-7pt}  
\begin{minipage}{0.50\linewidth} 
\vspace*{-7pt} 
  {\small
    \begin{eqnarray}
    \label{s-i-0-1}
    \mathtt{Human} & \sqsubseteq_n &
    \mathtt{\exists has\_heart. LH}
    \\
    \label{s-i-0-2}
    \mathtt{SI} & \sqsubseteq\, &
    \mathtt{Human}
    \end{eqnarray}}
\end{minipage}
\begin{minipage}{0.50\linewidth}  
\vspace*{-7pt}
  {\small
    \begin{eqnarray}
    \label{s-i-0-3}
    \mathtt{SI} & \sqsubseteq\, &
    \mathtt{\exists has\_heart. RH}
    \\
    \label{s-i-0-4}
    \mathtt{\exists has\_heart. LH} 
    &\sqsubseteq\, & \neg \mathtt{\exists has\_heart. RH}
    \end{eqnarray}}
\end{minipage}
where  $\mathtt{LH}$ (resp. $\mathtt{RH}$) denotes left-positioned (resp. right-positioned) hearts and $\mathtt{SI}$ stands for situs inversus. 

Since $\tmS\subseteq \KB^\Sigma$, by \eqref{s-i-0-3} and \eqref{s-i-0-4} we have that the instances of $\mathtt{SI}$ have their heart on the right-hand side (and hence not on the left-hand side):\\
\vspace*{5pt}
\begin{minipage}{0.50\linewidth}  
\vspace*{-10pt} 
\begin{eqnarray}
\label{situs-inversus-property}
  \KB&\nent&
    \mathtt{SI \sqsubseteq \exists has\_heart. RH}
\end{eqnarray}      
\end{minipage}
\begin{minipage}{0.50\linewidth}  
\vspace*{-10pt}
\begin{eqnarray}
\label{situs-inversus-property}
      \KB&\nent&
    \mathtt{SI \sqsubseteq \neg \exists has\_heart. LH} \,. 
\end{eqnarray}      
\end{minipage}
      
Then, let $\Sigma=\{\mathtt{\N Human}\}$. It is straightforward to see that $\KB^\Sigma$ consists of the strong axioms  \eqref{s-i-0-2} -- \eqref{s-i-0-4}, plus\\
\begin{minipage}{0.50\linewidth} 
{\small
    \begin{eqnarray}
    \mathtt{\N Human} & \sqsubseteq & \mathtt{Human}
    \end{eqnarray}
    }
\end{minipage}
\begin{minipage}{0.50\linewidth} 
{\small
    \begin{eqnarray}
    \mathtt{Human}\sqcap \mathtt{\N Human} & \sqsubseteq &
    \mathtt{\exists has\_heart. LH\,.}
    \end{eqnarray}
    }
\end{minipage}    
Consequently, we have that
\begin{equation}
  \KB\nent
    \mathtt{\N Human \sqsubseteq \exists has\_heart. LH} \,.
\end{equation} 

\noindent
Moreover, as a classical consequence of the above inferences, one
can further conclude that people with situs inversus are not
  standard humans:
{\small  
  \begin{equation}
    \label{inversus-not-normal}
    \KB \nent \mathtt{SI \sqsubseteq \neg \N Human } \,.
  \end{equation}  
}
\noindent
Conversely, if $\Sigma=\{\mathtt{\N SI}\}$, the iterative construction of $\KB^\Sigma$ adds in first step  the axiom  
{\small
    \begin{eqnarray}
    \label{situs-inversus-property}
    \mathtt{\N SI} & \sqsubseteq & \mathtt{SI}\, ,
    \end{eqnarray}
    } 
then the DI \eqref{s-i-0-1} is overridden, since adding 
{\small
    \begin{eqnarray}
    \mathtt{Human}\sqcap \mathtt{\N SI} & \sqsubseteq &
    \mathtt{\exists has\_heart. LH}
    \end{eqnarray}
    } 
    
\noindent    
would make, together with axioms \eqref{s-i-0-2}, \eqref{s-i-0-3},  \eqref{s-i-0-4}, and \eqref{situs-inversus-property}, $\mathtt{\N SI}$ inconsistent. Consequently, 
$\mathtt{\N SI}$ is simply a consistent subclass of 
$\mathtt{SI}$ that does not satisfy any further property.

Now, extend $\KB$ with the additional DI:
{\small
  \begin{equation}
    \label{s-i-01-0}
    \mathtt{Human} \sqsubseteq_n \mathtt{\exists has\_organ.Nose} \,.
  \end{equation}
}  
 Note that  \eqref{s-i-01-0} and \eqref{s-i-0-1} have both  maximal priority (indeed, they are incomparable by specificity). 
  It is easy to see that (\ref{s-i-01-0}) is
  overridden neither in $\N\mathtt{Human}$ nor in
  $\N\mathtt{SI}$, therefore both of the following
  inferences are valid:\\  
  \vspace*{-11pt}
{\small  
  \begin{eqnarray}
    \KB  &\nent&  \N\mathtt{Human} \sqsubseteq \mathtt{\exists
      has\_organ.Nose}\\ 
    \KB &\nent&  \N\mathtt{SI} \sqsubseteq \mathtt{\exists
      has\_organ.Nose} \,.
  \end{eqnarray}
  }
  In other words, the property of having a nose is inherited even if (\ref{inversus-not-normal}) make
  $\mathtt{SI}$ exceptional
  w.r.t.\ $\mathtt{Human}$. \qed
\end{example}

\begin{example}
  \label{nixon}
  
  \label{nixon-0.1}
  Consider the following variant of Nixon's diamond \cite{DBLP:journals/ai/BonattiFPS15}:\\
\begin{minipage}{0.45\linewidth}
{\small
  \begin{eqnarray}
    \label{nix-1-0}
    \mathtt{Quaker} & \sqsubseteq_n & \mathtt{Pacifist} \,,
    \\
    \label{nix-1-1}
    \mathtt{Republican} & \sqsubseteq_n & \neg \mathtt{Pacifist} \,,
  \end{eqnarray}
}  
\end{minipage}  
\begin{minipage}{0.55\linewidth}
{\small
  \begin{eqnarray}
    \label{nix-1-2}
    \mathtt{RepQuaker} & \sqsubseteq & \mathtt{Republican}
    \sqcap \mathtt{Quaker} \,.
  \end{eqnarray}
}  
\end{minipage}    

\noindent
  Note that  the two DIs (\ref{nix-1-0}) and (\ref{nix-1-1}) are incomparable under specificity. Moreover, since they  both  can be
  individually satisfied by $\N\mathtt{RepQuaker}$, without
  making it inconsistent, none of them is overridden in
  $\mathtt{\N RepQuaker}$.
  It follows that $\mathtt{\N RepQuaker}$ must satisfy both
  DIs and consequently $\mathtt{RepQuaker}$ is
  associated to an inconsistent prototype:
{\small  
  \begin{equation}\label{es:inc}
  \KB \nent \mathtt{\N RepQuaker}
  \sqsubseteq \bot\,.
  \end{equation} 
}
Note that even if the prototype of $\mathtt{RepQuaker}$ is
  inconsistent, the knowledge base is consistent, as well as many
  normality concepts.  In particular, we have $\KB \not\nent \N
  \mathtt{Quaker} \sqsubseteq \bot$ and $\KB \not\nent \N
  \mathtt{Republican} \sqsubseteq \bot$.
  
Moreover, consequence \eqref{es:inc} cannot be resolved by logic since (\ref{nix-1-0}) and (\ref{nix-1-1}) are perfectly symmetric w.r.t. $\N\mathtt{RepQuaker}$. 
 Removing the inconsistency is up to the knowledge engineer which has to choose how to repair $\KB$.
For instance,  adding $\mathtt{RepQuaker \sqsubseteq_n  Pacifist}$ 
  (resp. $\mathtt{RepQuaker \sqsubseteq_n \neg Pacifist}$)  resolves the conflict in favor of the first (resp. second) DI. 
\qed
\end{example}

\begin{remark}
  \label{rem:nixon}
  The other nonmonotonic semantics of DLs silently ``hide'' unresolved conflicts, by deriving none of the conflicting properties.  The result is a gap in the knowledge base. For example, Nixon was notoriously not a pacifist, and until the conflict is resolved, this information is not accessible to reasoners. In other examples, such knowledge gaps may have important consequences \cite{DBLP:journals/ai/BonattiFPS15}. Unlike the other nonmonotonic logics, \emph{\DLN helps knowledge engineers in identifying the gaps caused by unresolved conflicts}. Searching for inconsistent prototypes is analogous to the classical KB debugging activity consisting in identifying inconsistent concepts, and all engines support it. 
\end{remark}

\begin{example}
\label{ex:reservist}
In several countries (e.g. Mexico, Norway and Brazil) military service is mandatory for male citizens (except for special cases such as mental disorders). After military training, citizens become \emph{reservists}, and shall join the army again in case of war. This
can be formalized with the following DIs: {
    \begin{eqnarray}
      \label{ms1}
      & \mathtt{MaleCitizen}  \sqsubseteq_n 
      \mathtt{HasMilitaryTraining} &
      \\
      \label{ms2}
      & \mathtt{MaleCitizen\sqcap HasMilitaryTraining}  \sqsubseteq_n 
      \mathtt{Reservist}\,. &
    \end{eqnarray}
}%
The exceptions to the above rules include minors:
{
    \begin{eqnarray}
    \label{ms3}
    \mathtt{MinorMaleCitizen} & \sqsubseteq &
    \mathtt{MaleCitizen}
    \\
    \label{ms4}
    \mathtt{MinorMaleCitizen} & \sqsubseteq &
    \mathtt{\neg HasMilitaryTraining}\,.
    \end{eqnarray}
}%
Axiom (\ref{ms4}) should prevent (\ref{ms2}) from being applied to
minors, that is, it should \emph{not} be possible to conclude that
$\N\,\mathtt{MinorMaleCitizen} \sqsubseteq \mathtt{Reservist}$ (indeed,
this is what happens with \DLN).
\qed  
\end{example}

For what concerns the computational complexity of \DLN, notice that the iterative construction of  $\KB^\Sigma$ requires, at each step $i\in\{1,\ldots, \vert\D\vert\}$, (i) to restrict 
$\tmS^\Sigma_i$ to the  DIs that have a higher priority than $\delta_i$, and (ii) to evaluate $\vert \Sigma\vert$ consistency checks. If the priority relation is based on specificity, checking whether $\delta_j\prec \delta_i$ consists in solving two subsumption problems and hence it has the same complexity as entailment in \DL.   
Also the second point simply comes down to classical reasoning in the underlying $\DL$.  Consequently, $\DLN$ entailment has the same complexity as in $\DL$. In general, considering that different priority relations can be used, the following characterization holds.    
\begin{theorem}
  \label{thm:complx-0}
  Let \DL be a DL fragment such that subsumption (resp.\ instance) checking in \DL belongs to a complexity class \C, and deciding the preference relation $\prec$ belongs to $\PT^\C$.\footnote{$\PT^\C$ is the class of all problems that can  be solved by a deterministic Turing machine in polynomial time
    using an oracle for \C.}
  If \DL supports $\sqcap$ in the left-hand side of inclusions,
  then subsumption (resp.\ instance) checking in \DLN is in
  $\PT^\C$.
\end{theorem}

\noindent
Since $\PT^\PT$ equals \PT,  the entailment problem $\KB \nent \alpha$ is tractable in low complexity description logics such as   $(\mbox{\it DL-lite}_{horn}^\mathcal{(HN)})^\N$
\hspace{2pt}\cite{DBLP:journals/jair/ArtaleCKZ09} and (\mbox{\ELP$)^\N$}. 
Similarly, Theorem~\ref{thm:complx-0} tells us that $\mathcal{SROIQ}^\N$ reasoning is in $\PT^\mathrm{N2ExpTime}$ for suitable priority relations.


\section{Features and comparisons}\label{sec:feat}

\subsection{Knowledge engineering requirements}\label{sec:req}

The \DLN family of logics results from a utilitarian way of
approaching nonmonotonic logic design. The main goal of this approach is addressing
the practical needs of ontology and policy designers, that have been
illustrated with several examples in the literature on biomedical
ontologies and semantic web policies. Here is a summary of the main
shortcomings addressed by \DLN (see \cite{DBLP:journals/ai/BonattiFPS15} for more details and explanations):

\begin{itemize}
\item \emph{Inheritance blocking}. Most of the logics
  grounded on preferential semantics and rational closure block
  the inheritance of \emph{all} default properties towards exceptional
  subclasses (as opposed to overriding only the properties that are modified in those subclasses). \DLN's overriding mechanism does not suffer from this drawback (see Example~\ref{situs-inversus-00.0}).

\item \emph{Undesired CWA effects}. Many nonmonotonic DLs extend default
  properties to as many individuals as possible, thereby introducing CWA (i.e.\ closed-world assumption)
  effects that clash with the intended behavior of 
  ontologies. For instance, exceptional concepts (such as 
  $\mathtt{SI}$ in Example~\ref{situs-inversus-00.0}) collapse to the list of constants that are explicitly asserted to be in the concept (and if no such constants exist, exceptional concepts become inconsistent).  \DLN does not introduce any CWA
  effect because it does not force individuals to be normal, unless
  explicitly stated otherwise.

\item \emph{Control on priorities}. Since priorities are not fixed a
  priori in \DLN, knowledge engineers can adapt them to their needs.
  In principle, it is possible to override DIs based on temporal
  criteria (which may be useful in legal ontologies and ontology
  versioning), define default conflict resolution criteria,
  and even use rational closure's specificity-based axiom ranking. The logics derived from
  inheritance networks, preferential semantics, and rational closure
  can only support their fixed, specificity-based overriding
  criterion.  

\item \emph{Default role fillers}. Should role values be restricted to
  normal individuals? Sometimes, this kind of inference is desirable,
  sometimes it is not, cf.~\cite{DBLP:journals/ai/BonattiFPS15}. Some logics are completely
  unable to apply default properties to role values.\footnote{This is the case for rational closure. Recently, in \cite{Pensel-thesis2019}, a solution has been proposed for \EL with $\bot$. It is unclear how to extend it to more expressive DLs, and it is not possible to ``turn off'' the application of default rules to role fillers.}  Some others
  cannot switch this inference off when it is not desired.  Only \DLN
  and $\ALC+\ty_{\min}$ make it possible to control this kind of
  inference. Due to its explicit priorities, \DLN is also able to
  encode a design pattern that makes role ranges normal whenever this
  does not override any explicit DI.

\item \emph{Inconsistent prototype detection}. We argued that
  when conflicts cannot be settled by priorities, silent conflict resolution
  is not a desirable feature: knowledge engineers should be involved
  because there is no universally correct automated resolution
  criterion (cf.\ Example~\ref{nixon}). Only \DLN
  and probabilistic description logics (and $\ALC+\ty_{\min}$, in some
  very specific cases) detect inconsistent prototypes and make them
  evident, as advocated in Remark~\ref{rem:nixon}.

\item \emph{Unique deductive closure}. As a result of automated
  conflict resolution, several nonmonotonic logics yield multiple
  deductive closures, corresponding to all the alternative ways of
  solving each conflict. \DLN is one of the logics that has a unique
  closure.

\item \emph{Generality}. Nonmonotonic extensions should be applicable to all description logics, or at least to the standard OWL2-DL (i.e.\ the logic ${\cal SROIQ}$(\D)).  Typicality logics and rational closure, instead, are limited to logics that satisfy the \emph{disjoint union model property}. Recently, it has been shown that for expressive DLs that do not enjoy this property, syntactic inference does not match semantics \cite{DBLP:journals/ai/Bonatti19}.
  The same paper introduces \emph{stable rational closure} that solves the generality problem for rational closure, but re-introduces the issue of multiple (or non existent) deductive closures. It is currently not clear how to design a logic that satisfies the KLM postulates, is fully general, and yields a unique closure for all knowledge bases.

\item \emph{Low complexity}. \DLN
  preserves the tractability of these reasoning tasks for all
  low-complexity DLs, including the rich tractable logics \ELP and
  DL-lite$_\mathit{Horn}^\mathcal{(HN)}$.  Currently, no other
  nonmonotonic DL enjoys this property to the same extent.
  Rational closure has been proved to be tractable for \EL extended with $\bot$ 
  \cite{DBLP:journals/isci/CasiniSM19,Pensel-thesis2019}.
  Some logics, such as
  \cite{DBLP:conf/jelia/CasiniS10,DBLP:conf/dlog/CasiniMMV13,DBLP:journals/jair/CasiniS13,DBLP:journals/fuin/GiordanoOGP09,DBLP:journals/ai/GiordanoGOP13}, preserve
  the asymptotic complexity of ExpTime-complete DLs like \ALC.  More
  generally, \DLN preserves the asymptotic complexity of all the DLs
  that belong to a deterministic complexity class that contains P. For nondeterministic complexity classes \C, an upper bound is $P^\C$.
 \end{itemize}

\DLN has been designed to address the above practical issues. In general, the utilitarian approach led us
to make \DLN neutral with respect to the inferences that are not
always desired: when possible, \DLN gives knowledge engineers the
ability of switching those inferences on and off.  The final result of
this investigation is a logic that enjoys a unique set of
properties, as shown by the summary in Table~\ref{summary-of-comparisons}.

\begin{table}
  \caption{Summary of comparisons with nonmonotonic DLs}
  \label{summary-of-comparisons}
  \footnotesize
  \begin{center}
    \begin{tabular}{|m{8.5em}||>{\centering\arraybackslash}m{3.11em}|>{\centering\arraybackslash}m{2.2em}|>{\centering\arraybackslash}m{2.1em}|>{\centering\arraybackslash}m{3.11em}|>{\centering\arraybackslash}m{3em}|>{\centering\arraybackslash}m{1.6em}|>{\centering\arraybackslash}m{1.9em}|>{\centering\arraybackslash}m{2em}|}
      \hline
      \multicolumn{1}{|p{8.5em}||}{} &
      \multicolumn{1}{c|}{CIRC} &
      \multicolumn{1}{c|}{DEF} &
      \multicolumn{1}{c|}{AEL} &
      \multicolumn{1}{c|}{TYP} &
      \multicolumn{2}{c|}{RAT} &
      \multicolumn{1}{c|}{PR} &
      \\
      \textbf{Features}
      & \footnotesize 
        \cite{DBLP:journals/jair/BonattiLW09,DBLP:journals/jair/BonattiFS11}
      & \footnotesize 
        \cite{DBLP:journals/jar/BaaderH95,DBLP:journals/jar/BaaderH95a}
      & \footnotesize 
        \cite{DBLP:journals/tocl/DoniniNR02}
      & \footnotesize 
        \cite{DBLP:journals/fuin/GiordanoOGP09,DBLP:journals/ai/GiordanoGOP13}
      & 
        \multicolumn{1}{c}{
        \cite{DBLP:conf/jelia/CasiniS10,DBLP:conf/dlog/CasiniMMV13}
        }
      & \footnotesize
        \cite{DBLP:journals/jair/CasiniS13} 
      & \footnotesize 
        \cite{DBLP:journals/ai/Lukasiewicz08} 
      & \footnotesize 
        \DLN 
      \\
      \hline
      \hline
      no inheritance~~~~~\break blocking
      & \checkmark 
      & \checkmark 
      & \checkmark 
      &  
      &  
      & \checkmark 
      & \checkmark 
      & \checkmark 
      \\
      \hline
      no CWA effects
      &  
      & \checkmark 
      & \checkmark 
      &  
      & \checkmark 
      & \checkmark 
      &  
      & \checkmark 
      \\
      \hline
      fine-grained control~\break on role ranges
      &  
      &  
      &  
      &  ~\checkmark$^{(1)}$ 
      &  
      &  
      &  
      & \checkmark 
      \\
      \hline
      detects inconsistent~\break prototypes
      &  
      &  
      &  
      & ~\checkmark$^{(1)}$ 
      &  
      &  
      & ~\checkmark$^{(2)}$ 
      & \checkmark 
      \\
      \hline
      unique deductive~~~\break closure
      & \checkmark 
      &  
      &  
      & \checkmark 
      &  
      &  
      &  
      & \checkmark 
      \\
\hide{      \hline
      preserves legacy~~~\break taxonomies
      &  
      &  
      &  
      &  
      &  
      &  
      &  
      & \checkmark 
      \\
}
      \hline
      preserves tractability
      &  
      &  
      &  
      &  
      &  
      &  
      &  
      & ~\checkmark 
      \\
      \hline
      generality
      & \checkmark 
      & \checkmark 
      & \checkmark 
      &  
      &  
      &  
      & \checkmark 
      & ~\checkmark 
      \\
      \hline
      \hline
      implicit specificity
      &  
      &  
      &  
      & \checkmark 
      & \checkmark 
      & \checkmark 
      & \checkmark 
      &  
      \\
      \hline
      other priorities
      & \checkmark 
      & \checkmark 
      &  
      &  
      &  
      &  
      &  
      & \checkmark 
      \\
      \hline
      \multicolumn{9}{l}{\footnotesize (1)
        \vphantom{$A^{B^C}$}
        Partially supported.}
      \\
      \multicolumn{9}{l}{\footnotesize (2)\,
        Inconsistency may propagate to the entire KB.}
      \\
    \end{tabular}
  \end{center}
\end{table}

We deliberately refrained from
adding a priori any requirements that are not directly motivated by
applications, such as the KLM postulates. Interestingly, \DLN satisfies many of those postulates, though, as illustrated in the next section.

\subsection{\DLN and the KLM axioms}\label{sec:klm}

\begin{table}[h]
  \caption{The KLM postulates in \DLN}
  \label{tab:rational-closure}
\begin{center}
  \renewcommand{\arraystretch}{2}
  {\small 
  \begin{tabular}{ccc}
 
    \hline
    Name & Rule schema & Sound in \DLN
    \\
    \hline
    REF & \infrule{\alpha\in\KB}{\KB \nent \alpha} & \checkmark
    \\
    CT & \infrule{\KB\nent \alpha \quad \KB \cup\{\alpha\}\nent \gamma}{\KB \nent \gamma} & \checkmark
    \\
    CM & \infrule{\KB \nent \alpha \quad \KB\nent\gamma}{\KB \cup \{\alpha\} \nent \gamma} & \checkmark
    \\
    LLE & \infrule{\KB \cup \{\alpha\} \nent \gamma \quad \models \alpha\equiv \beta}{\KB \cup \{\beta\}\nent \gamma} & \checkmark
    \\
    RW & \infrule{\KB \nent \alpha \quad \alpha \models \gamma}{\KB \nent \gamma} & \checkmark
    \\
    OR & \infrule{\KB \cup \{\alpha\} \nent \gamma \quad \KB \cup \{\beta\} \nent \gamma}{\KB \cup \{\alpha \lor \beta\}\nent \gamma} & under extra axioms
    \\
    RM & \infrule{\KB \nent \gamma \quad \KB\not\nent\neg\alpha}{\KB \cup \{\alpha\} \nent \gamma} & under extra axioms
    \vspace*{.1em}
    \\
    \hline
    \multicolumn{3}{l}{\footnotesize \KB is a canonical \DLN knowledge base;}
    \vspace*{-1.2em}
    \\
    \multicolumn{3}{l}{\footnotesize $\alpha$ and $\beta$ range over \DL assertions and (strong) concept and role inclusions;}
    \vspace*{-1.2em}
    \\
    \multicolumn{3}{l}{\footnotesize $\gamma$ ranges over \DLN assertions and \DLN concept/role inclusions;}
    \vspace*{-1.2em}
    \\
    \multicolumn{3}{l}{\footnotesize nonstandard DL axioms $\alpha\lor\beta$, $\neg \beta$ can be simulated, e.g.\ with the universal role;}
    \vspace*{-1.2em}
    \\
    \multicolumn{3}{l}{\footnotesize \nent denotes the nonmonotonic consequence relation of \DLN and $\models$  classical inference.}
    \\
    \hline
  \end{tabular}
  }
\end{center}
\end{table}


In this section we analyze the logical properties of \DLN through the KLM postulates. 
In  \cite{DBLP:journals/ai/KrausLM90,DBLP:journals/ai/LehmannM92,DBLP:journals/amai/Lehmann95}, Kraus, Lehmann, and Magidor argued that in order to reason about what
normally holds in the world, it is desirable to make nonmonotonic
consequence relations closed under certain properties, called
\emph{KLM postulates}. 
Although these postulates are not necessarily desiderata,
due to the loose correspondence between their motivations
and \DLN's goals and semantics (cf. \cite{DBLP:journals/ai/BonattiFPS15,DBLP:journals/ai/BonattiS17}), we regard them as a useful technical tool for comparison, since the validity of
the postulates has been extensively investigated in most nonmonotonic logics.


There exist several versions of the postulates; all of them contain postulates that are incompatible with \DLN's novel way of highlighting unresolved conflicts through inconsistent prototypes, for debugging purposes. So, hereafter, \emph{we assume that all unresolved conflicts have been fixed}, as recommended by this knowledge engineering methodology.

The first version of the postulates -- illustrated in
Table~\ref{tab:rational-closure} -- is the verbatim instantiation of
the original, meta-level postulates. A consequence relation that
satisfies the KLM postulates is called \emph{rational}.  It is called
\emph{preferential} if it satisfies all rules but RM, and
\emph{cumulative} if it satisfies all rules but RM and OR. With
respect to this version of the postulates, \DLN's consequence relation
($\nent$) is \emph{cumulative}.\footnote{Obviously, REF and RW always hold,
  because \DLN is closed under classical inference. Moreover rules CT,
  CM, and LLE are sound by \cite[Thm~1]{DBLP:journals/ai/BonattiS17}.}

%

It is interesting to note that the rational closure of DLs itself is \emph{not} rational w.r.t.\ this version of the postulates, e.g.\ it fails to satisfy the OR rule \cite{DBLP:journals/ai/BonattiS17}.
On the contrary, \DLN can be made fully rational by making it a little more similar to typicality logic and rational closure, in the following respect.  The semantic of typicality and rational closure forces each consistent concept to have a normal instance, through the \emph{smoothness} property of preferential models and the notion of \emph{canonical model} \cite{DBLP:journals/ai/GiordanoGOP15}. A similar condition can be enforced in \DLN through the axioms $\neg(\N C\sqsubseteq\bot)$, for all consistent $C$ occurring in \KB or in the query.  The knowledge bases so extended \emph{satisfy all the postulates in Table~\ref{tab:rational-closure}.}
The above results provide an immediate comparison with the consequence relations of Circumscribed DLs (that are preferential) and those of Default and Autoepistemic DLs (that are not cumulative).

\begin{table}
\centering
\makebox[0pt][c]{\parbox{1\textwidth}{%
    \begin{minipage}[b]{0.45\hsize}\centering
      \renewcommand{\arraystretch}{2}
        \begin{tabular}{ccr}
    \hline
    Name & Rule schema & Sound 
    \\
    \hline
    REF$_n$ & \infrule{~}{C\sqsubseteq_n C} & \checkmark
    \\
    CT$_n$ & \infrule{C\sqsubseteq_n D \quad C\sqcap D\sqsubseteq_n E}{C\sqsubseteq_n E} & 
    \\
    CM$_n$ & \infrule{C\sqsubseteq_n D \quad C\sqsubseteq_n E}{C\sqcap D\sqsubseteq_n E} &
    \\
    LLE$_n$ & \infrule{C\sqsubseteq_n E \quad \tmS \models C\equiv D}{D\sqsubseteq_n E} & 
    \\
    RW$_n$ & \infrule{C\sqsubseteq_n D \quad \tmS \models D\sqsubseteq E}{C\sqsubseteq_n E} &
    \\
    OR$_n$ & \infrule{C\sqsubseteq_n E \quad D\sqsubseteq_n E}{C\sqcup D\sqsubseteq_n E} & 
    \\
    RM$_n$ & \infrule{C\sqsubseteq_n E \quad C\not\sqsubseteq_n \neg D}{C\sqcap D\sqsubseteq_n E} &
   \vspace*{.5em}
    \\
    \hline
    \multicolumn{3}{l}{\footnotesize \tmS is the strong part of a knowledge base}
    \\
    \hline
  \end{tabular}
     \vspace*{.5em}
     \caption{Analogues of the KLM postulates for DIs \cite{DBLP:journals/ai/BonattiFPS15}}
     \label{tab:DI-KLM}
    \end{minipage}
    \hfill
    \begin{minipage}[b]{0.45\hsize}\centering
      \renewcommand{\arraystretch}{2}
         \begin{tabular}{ccr}
    \hline
    Name & Rule schema & Sound 
    \\
    \hline
    REF$^N$ & \infrule{~}{\N C\sqsubseteq C} & \checkmark
    \\
    CT$^N$ & \infrule{\N C\sqsubseteq D \quad \N(C\sqcap D)\sqsubseteq E}{\N C\sqsubseteq E} & partly
    \\
    CM$^N$ & \infrule{\N C\sqsubseteq D \quad \N C\sqsubseteq E}{\N(C\sqcap D)\sqsubseteq E} & partly
    \\
    LLE$^N$ & \infrule{\N C\sqsubseteq E \quad C\equiv D}{\N D\sqsubseteq E} &
    \\
    RW$^N$ & \infrule{\N C\sqsubseteq D \quad D\sqsubseteq E}{\N C\sqsubseteq E} & \checkmark
    \\
    OR$^N$ & \infrule{\N C\sqsubseteq E \quad \N D\sqsubseteq E}{\N(C\sqcup D)\sqsubseteq E} & partly
    \\
    RM$^N$ & \infrule{\N C\sqsubseteq E \quad \N C\not\sqsubseteq \neg D}{\N(C\sqcap D)\sqsubseteq E} & partly
    \vspace*{.5em}
    \\
    \hline
    \multicolumn{3}{l}{\footnotesize $C$, $D$, and $E$ range over \DL concepts}
    \\
    \hline
  \end{tabular}
  \vspace*{.5em}
  \caption{\DLN Candidate inference rules inspired by KLM postulates}
  \label{tab:N-KLM}
    \end{minipage}%
}}
\end{table}

Several logics internalize the nonmonotonic consequence
relation and push the KLM postulates to the object level
(e.g. \cite{DBLP:conf/jelia/CasiniS10,DBLP:conf/ijcai/CasiniS11,DBLP:journals/jair/CasiniS13,DBLP:conf/jelia/CasiniMMN14,DBLP:conf/ausai/BritzMV11,DBLP:journals/ai/GiordanoGOP13,DBLP:journals/ai/GiordanoGOP15}). 
The resulting postulates for \DLN are reported in Table~\ref{tab:DI-KLM}. 
Their validity is clearly affected by overriding. For instance, if the second premise of
CM$_n$ were overridden, then there would be no logical ground for supporting the conclusion. 
However, if the premises are not overridden, then all the postulates of Table~\ref{tab:DI-KLM} are valid in
\DLN. 
It is interesting to note that a similar phenomenon can be observed in
Lehmanns account of default reasoning \cite{DBLP:journals/amai/Lehmann95}. In Sec. 6 Lehmann exhibits a
knowledge base with no (consistent) rational closure; however it has a
lexicographic closure because the latter ignores all overridden defaults.
%
%
The second interesting remark is that two of these postulates unconditionally hold 
in most practically interesting cases: (i) the OR$_n$ rule holds if the priority relation is specificity; 
(ii)  LLE$_n$ holds whenever the priority relation is not sensitive to syntactic details i.e. 
treats logically equivalent DIs in the same way (like specificity does).

Another internalized version of the postulates,
analogous to those satisfied by typicality logics \cite{DBLP:journals/ai/GiordanoGOP13,DBLP:journals/ai/GiordanoGOP15} is reported in Table~\ref{tab:N-KLM}.
It can be shown that typicality DLs satisfy these postulates 
only because the normality criterion is assumed to be concept-independent (i.e. if John is
more typical than Mary as a driver, then he must also be more typical than Mary as a worker, 
as a tax payer, and so on) \cite{DBLP:journals/ai/BonattiS17}. \DLN does not adopt this strong assumption: 
in \DLN each concept may have its own notion of what is more normal or standard and -- for this reason -- 
it does not universally satisfy CT$^N$ and CM$^N$. 
So an interesting open question is whether the postulates of Table~\ref{tab:N-KLM} 
can possibly be satisfied by a logic that does not rely on a unique, concept-independent normality relation.

%
%
%

Again, \DLN can be made fully rational (with respect to this version
of the postulates) by making it more similar to rational
closure. Consider the restriction of \DLN where N does not explicitly
occur in KB (N can be used only in queries).  In practice, this means
that role fillers cannot be forced to be normal, similarly to what
inevitably happens in rational closure and default DLs, due to the
limitations of these logics.  Under this restriction, \emph{all}
postulates in Table~\ref{tab:N-KLM} hold \cite[Thm~4,~5,~6]{DBLP:journals/ai/BonattiS17}.
%
%
%

\section{Conclusions and future work}\label{sec:concl}

\DLN addresses a number of drawbacks that affect the nonmonotonic
semantics of DLs.  It has been designed with a particular attention to
practical  issues that hinder the adoption of nonmonotonic semantics in
OWL2 and its profiles, including expressiveness limitations, and
complexity problems. Moreover, \DLN reasoning can be easily reduced to
classical reasoning, thereby leveraging the high-quality,
well-engineered implementations of DL reasoning. In the light of these properties, \DLN compares favorably to the other nonmonotonic DLs, as summarized in Table~\ref{summary-of-comparisons}.

Interestingly, even if KLM postulates played no role in \DLN's design, \DLN satisfies to a large extent the major meta-level and internalized versions of the postulates. The postulates in Table~\ref{tab:DI-KLM} hold up to overriding, as in Lehmann's lexicographic closure. Moreover, \DLN is flexible enough to satisfy \emph{all} the other postulates by means of additional axioms or syntactic restrictions that make \DLN more similar to typicality logics and rational closure.

Many different directions deserve further investigation.
From a semantic viewpoint, we mentioned  in Section~\ref{sec:klm} that  the KLM postulates OR and RM are sound in \DLN under the assumption that consistent concepts have at least one typical individual. 
Taking inspiration from typicality logics, 
variants of \DLN can be investigated where this assumption is
hard-coded in the semantics. This should be done with some care,
though: if too many individuals were forced to be normal, then the
undesirable closed-world assumption effects described in
\cite{DBLP:journals/ai/BonattiFPS15}, that affect typicality logics,
might be introduced in \DLN.

The study of the logical properties of \DLN can be refined by
investigating the mutual relationships between DIs and their effects
on normality concepts. In particular, it would be interesting to
investigate hybrid versions of the postulates, whose premises are
taken from Table~\ref{tab:DI-KLM} while consequents are taken 
from Table~\ref{tab:N-KLM}.

Finally, one of the primary strengths of \DLN is that it preserves the tractability of the low-complexity DLs 
underlying the OWL2-EL and OWL2-QL profiles. However, asymptotic tractability alone does not suffice for practical purposes. In \cite{DBLP:conf/semweb/BonattiPS15}, two optimization techniques have been successfully applied to obtain real-time query answering over large knowledge bases.  One optimization is based precisely on a suitably modified module extraction algorithm, that so far constitutes the most effective optimization technique for \DLN (excluding combined approaches).  The other optimization, called \emph{optimistic method},
reduces the number of retractions (an expensive class of operations in incremental reasoning).
\DLN's module extractor, however, proved to be less effective for KBs that contain many explicit occurrences of the
normality concepts, and for those with nonempty ABoxes (due to the lesser effectiveness of the underlying classical module extractors in such contexts). 
To overcome these problematic cases, we plan to improve the module extractor for \DLN by discarding the normality concepts (and related axioms) and assertions that are irrelevant to a given query.

\bibliography{nonmon-dl}
\bibliographystyle{abbrv}
 
\end{document}